\documentclass[10pt,twocolumn,letterpaper]{article}

\usepackage[pagenumbers]{cvpr} %

\usepackage{tabularx}
\usepackage{booktabs}
\usepackage{svg}
\usepackage{makecell}
\usepackage{array}
\usepackage{tikz}

\newcommand{\PAR}[1]{\vskip4pt \noindent {\bf #1~}}

\newcolumntype{Y}{>{\centering\arraybackslash}X}
\newcolumntype{L}[1]{>{\raggedright\arraybackslash}p{#1}}

\newcommand{\mytt}[1]{{\footnotesize\texttt{#1}}}

\newcommand{\graymidrule}{\arrayrulecolor{black!30}\midrule\arrayrulecolor{black}}

\newcommand{\downrightarrow}{\hspace{5pt}%
  \raisebox{1.5pt}{\begin{tikzpicture}[scale=0.4, baseline=(current bounding box.south)]
    \draw[-latex, rounded corners=2pt] (0.0, 0.62) -- (0.0, 0.25) -- (0.7, 0.25);
  \end{tikzpicture}} %
}

\definecolor{cvprblue}{rgb}{0.21,0.49,0.74}
\usepackage[pagebackref,breaklinks,colorlinks,allcolors=cvprblue]{hyperref}

\def\paperID{27} %
\def\confName{CVPR}
\def\confYear{2025}

\title{How Important are Videos for Training Video LLMs?}

\author{
George Lydakis$^1$ %
\quad
Alexander Hermans$^1$%
\quad
Ali Athar$^2$
\quad
Daan de Geus$^{1,3}$%
\quad
Bastian Leibe$^1$\\[5pt]
$^1\hspace{1pt}$RWTH Aachen University \quad $^2\hspace{1pt}$ByteDance Seed \quad $^3\hspace{1pt}$Eindhoven University of Technology\\[5pt]
{\tt\small \{lydakis,hermans,leibe\}@vision.rwth-aachen.de}
\quad {\tt\small ali.athar@bytedance.com}
\quad {\tt\small d.c.d.geus@tue.nl} \\
}

\begin{document}
\maketitle
\begin{abstract}
Research into Video Large Language Models (LLMs) has progressed rapidly, with numerous models and benchmarks emerging in just a few years.
Typically, these models are initialized with a pretrained text-only LLM and finetuned on both image- and video-caption datasets.
In this paper, we present findings indicating that Video LLMs are more capable of temporal reasoning after image-only training than one would assume, and that improvements from video-specific training are surprisingly small.
Specifically, we show that image-trained versions of two LLMs trained with the recent LongVU algorithm perform significantly above chance level on TVBench, a temporal reasoning benchmark.
Additionally, we introduce a simple finetuning scheme involving sequences of annotated images and questions targeting temporal capabilities.
This baseline results in temporal reasoning performance close to, and occasionally higher than, what is achieved by video-trained LLMs.
This suggests suboptimal utilization of rich temporal features found in real video by current models.
Our analysis motivates further research into the mechanisms that allow image-trained LLMs to perform temporal reasoning, as well as into the bottlenecks that render current video training schemes inefficient.
\end{abstract}
    
\section{Introduction}
\label{sec:intro}

Initially developed as text-oriented architectures, Large Language Models (LLMs) have by now proven to be competitive choices for multi-modal reasoning as well.
Video-language tasks are no exception: a significant number of Video LLMs have appeared in recent years \cite{lin2024videollava, xu2024pllava, cheng2024videollama2, li2024llamavid, bai2025qwen2, yuan2025tarsier2, li2025temporal, shen2024longvu, he2024ma}.
The typical setting here is to either query the model for a description of the video, or to request an answer to a specific question related to it.

The majority of works in this domain utilize the same fundamental architecture: video frames are encoded into tokens using a strong pretrained model such as DINOv2 \cite{oquab2023dinov2} or CLIP \cite{radford2021learning}, and a subset of these tokens is concatenated with the prompt text embeddings.
The LLM is then trained---either as a whole or with adaptation methods such as LoRA \cite{hu2022lora}---to predict the desired output, be it a description of the video or the answer to a question.
The exact scheme for video tokenization has been the subject of extensive research \cite{shen2024longvu, xu2024pllava, li2024llamavid, cheng2024videollama2, he2024ma}, and so have various aspects of the training data, from the size and variety of the video corpus, to the quality and level of detail of the annotations \cite{bai2025qwen2, wang2024tarsier, yuan2025tarsier2, li2024mvbench, chen2024sharegpt4video}.

Our research is motivated by the observation that current training procedures for Video LLMs often involve three stages \cite{shen2024longvu, xu2024pllava, lin2024videollava, liu2024oryx}, and that only the last stage involves actual video data: first text-based LLM training, followed by multi-modal finetuning on image-text datasets, and finally multi-modal finetuning on video-text datasets.
This leads us to ask one key question: \emph{what part of the temporal reasoning capabilities of Video LLMs comes from being trained on videos}, and what part is \emph{already there} from text and image pretraining? This could give insights into the effectiveness of existing architectures to learn from different types of data, and particularly the video-text pairs from current video datasets.

To shed some light on this, we assess the temporal reasoning capabilities of Video LLMs after several training stages.
For these experiments, we use TVBench~\cite{cores2025losttimenewtemporal}, a multiple-choice question answering (MCQA) benchmark for evaluating temporal aspects of video understanding.
Surprisingly, we find that the image-trained versions of two Video LLMs trained with the recent LongVU \cite{shen2024longvu} approach perform significantly above chance level on some of the benchmark's tasks almost out of the box, and without ever being trained on real videos.
Furthermore, we find that finetuning them on \emph{pseudo videos} formed by concatenating unrelated captioned images yields performances close to, and occasionally higher than, those achieved by video-trained counterparts.
This indicates a potential bottleneck in the utilization of video datasets by current Video LLMs, either in the quality of the data, in the architectures, or a combination thereof.

Our analysis raises interesting research questions.
For one, what mechanisms render image-trained LLMs capable of temporal reasoning?
For another, what factors make video-based training significantly less effective than expected?
Since training Video LLMs is computationally expensive, we hope to motivate research effort towards identifying these factors and proposing improvements.

\section{Related work}
\label{sec:related}

The pace of research in Video LLMs has been rapid, resulting in a large number of models being released in a short timeframe.
Works such as Video-ChatGPT \cite{maaz2024videochatgpt}, Video-LLaMA \cite{zhang2023video}, and Video-LLaVA \cite{lin2024videollava} introduced a basic architecture consisting of a text-pretrained LLM, image and/or video encoders that tokenize video input, and simple modules---typically a linear layer or small MLP---that map these video tokens to a feature space interpretable by the LLM.
While this architecture has largely remained the same, several video tokenization schemes have been proposed, often with the purpose of increasing the number of frames being processed.
These may involve pooling \cite{xu2024pllava, cheng2024videollama2}, static \cite{li2024llamavid} or adaptive compression schemes \cite{shen2024longvu}, and memory banks~\cite{he2024ma}.

\PAR{Training schemes.} 
An important line of work, particularly relevant to our study, focuses on the quantity and quality of the training data.
For example, Tarsier \cite{wang2024tarsier} demonstrates that the simple architecture used in most modern Video LLMs achieves strong performance when trained on a large, properly filtered dataset from both proprietary and public sources.
Prior to that, works such as ShareGPT4Video \cite{chen2024sharegpt4video} and VideoChat2 \cite{li2024mvbench} also underlined the importance of including a variety of data sources in terms of video content and question format.

A general trend in the literature is for Video LLMs to be trained on a combination of images and videos, either mixed \cite{lin2024videollava, liu2024oryx} or in two separate training phases \cite{shen2024longvu, xu2024pllava}.
For video training specifically, several methods focus on the creation of text annotations with desirable properties.
Examples include aligning the model's predictions with human-preferred output \cite{bai2025qwen2}, targeting both summarization as well as temporal localization \cite{li2025temporal}, augmenting captions with frame ranges that localize events \cite{yuan2025tarsier2} and automatically generating captions at different temporal granularities \cite{zhang2024llavavideo}.
So far, however, the relative importance of training on videos compared to text and images to enable temporal understanding capabilities has not been extensively studied.

\PAR{Evaluating Video LLMs.} 
In order to evaluate the temporal reasoning capabilities of Video LLMs, an appropriate benchmark is necessary.
Alongside model development, several evaluation datasets and methodologies have been proposed.
These can be classified into two broad categories, depending on the format of their questions.
The first is based on ``free-form'' questions, where the model is given a video and a question and is expected to produce an answer without any other cues.
Examples include MSRVTT \cite{xu2016msrvtt}, TGIF \cite{jang2017tgif} and ActivityNet-QA \cite{yu2019activitynet}.
Evaluation of the quality of answers is typically done with the use of another LLM which is tasked to assign a score to the output of the Video LLM based on the given ground truth answer.
The second category is based on multiple-choice questions, where the model's output is expected to match one of the given options.
Here we find benchmarks such as MVBench \cite{li2024mvbench}, Video-MME \cite{fu2024videomme}, TVBench \cite{cores2025losttimenewtemporal} and LongVideoBench \cite{wu2024longvideobench}.
In this case, the metric is the standard accuracy.

The main advantage of the first paradigm is that no effort is needed to design wrong options that are not trivial to reject, and that no hints are given to the Video LLM.
However, the metric is less interpretable and strongly dependent on the LLM used for evaluation.
Because (1) our study focuses specifically on the temporal reasoning capabilities of Video LLMs, and (2) we aim for our results to be independent of the choice of LLM, we opt for the multiple-choice TVBench benchmark\cite{cores2025losttimenewtemporal}. TVBench was specifically designed to address deficiencies of prior Video LLM benchmarks---namely, that many questions could be answered from a single frame or even from the text query alone.
By specifically designing both questions and distractors to require temporal understanding, TVBench alleviates this problem, making it well-suited for our study.

\section{Method}
\label{sec:method}

\begin{figure*}[th]
    \centering
    \includegraphics[width=\linewidth]{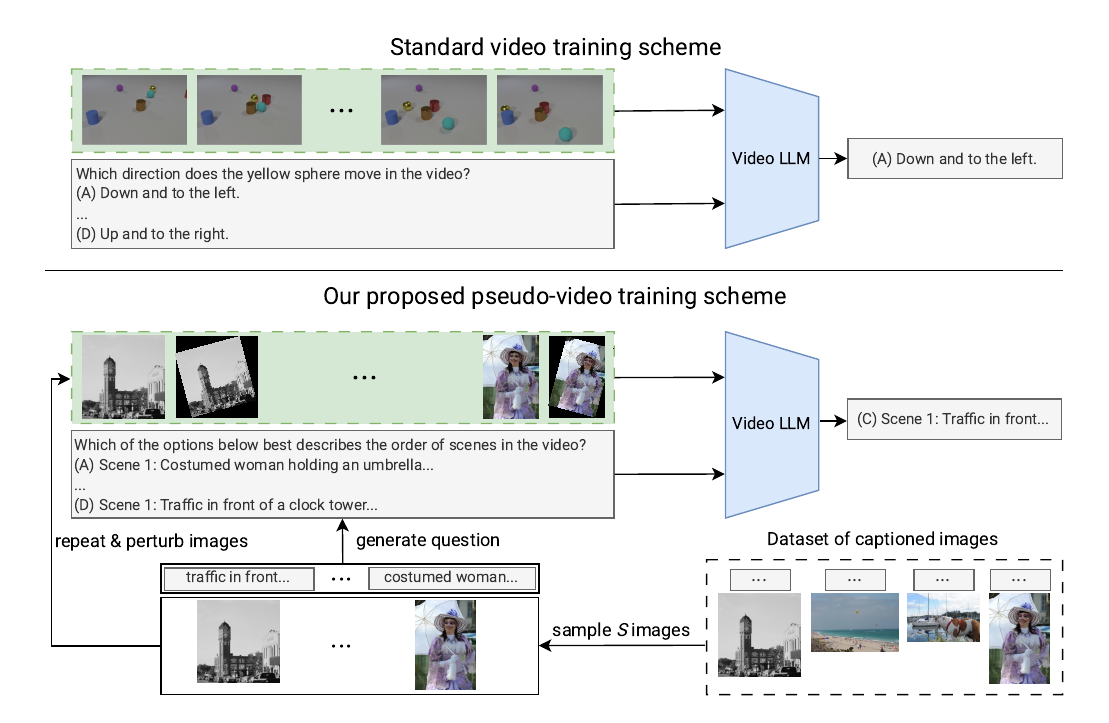}
    \caption{A comparison of the standard video-based training  scheme for Video LLMs (top) and our proposed pseudo-video training scheme (bottom). We utilize captioned image datasets to automatically generate short pseudo videos and questions for training.}
    \label{fig:video_llms}
\end{figure*}

A significant challenge in training Video LLMs is the fact that such models must possess reasoning capabilities of three distinct kinds, as well as the ability to combine all of them.
First, they must be able to parse text effectively and extract information about what is being asked.
Second, the questions that are posed are very diverse in terms of required visual features, which may need to encode \eg object types, colors, shapes, cardinalities and spatial positions.
Third, reasoning about videos requires understanding of the \emph{evolution} of these features along time.

The high complexity and compositionality of the task make apparent why these models are initialized with a text-pretrained LLM, and why they also utilize annotated image datasets during training.
In this work, we are interested in investigating the relative impact of the video datasets compared to that of text-image pretraining.
In other words, to what extent are Video LLMs acquiring temporal reasoning capabilities from their video finetuning?

To this end, we propose studying their temporal reasoning behavior in two settings that do not involve training on real videos.
First, we evaluate models that have only been trained on text and images.
Second, we substitute the video-training stage by the following procedure for utilizing image-caption datasets in temporal tasks.
Assuming a captioned image dataset $D = \{d_1, d_2, \ldots, d_N\}$ with $N$ samples, where $d_j = (i_j, c_j)$ is a pair consisting of an image $i_j$ and a corresponding caption $c_j$, we construct \emph{pseudo videos} as follows.
First, $S$ image-caption samples $d_j$ are randomly selected from the full dataset $D$, each of which forms the basis for a single video ``scene''.
Subsequently, a duration $F_j$ is selected for each of these scenes, and the corresponding image $i_j$ is repeated $F_j$ times.
A relatively mild affine transformation is applied to each frame, in order to simulate---at a very coarse level---a video.
Consequently, the resulting pseudo video has a length of $\sum_{j=1}^{S} F_j$ frames.

The advantage of this otherwise simple approach is that the temporal evolution of the pseudo video is fully known.
Assuming relatively noise-free image captions, we can thus generate various questions that require the model to reason about time.
Figure \ref{fig:video_llms} (bottom) illustrates the training process, compared to standard video training (Figure \ref{fig:video_llms} (top)).
Our intention here is for this to serve as a baseline: pseudo videos are significantly less information-rich than real videos, and less similar to real data seen during inference.
Therefore, a well-functioning video training scheme is expected to perform significantly better than this baseline.

The quality of learned representations may be influenced by factors such as the duration of individual scenes, their number and, perhaps more importantly, the questions posed to the model.
In this work, we study multiple-choice questions where exactly one answer is correct.
\begin{table*}[!t]
\centering
\small
\begin{tabularx}{\textwidth}{L{2cm}X}
\toprule
\textbf{Abbreviation} & \textbf{Question description} \\
\midrule
$\mathbf{R_1}$ & ``\mytt{Which of the following options best describes the order of scenes in the video?}'', followed by a number of different permutations of the scene captions, each of which has the form ``\mytt{Scene 1:} [caption] \mytt{Scene 2:} [caption] \dots \mytt{Scene} [$N$]\mytt{:} [caption]'' \\
\graymidrule
$\mathbf{R_2}$ & ``\mytt{In the given video, does the scene that can be captioned as "}[caption]\mytt{" happen before or after the scene that can be captioned as "}[caption]\mytt{"?}'', always followed by exactly two answer options, ``\mytt{before}'' and ``\mytt{after}''. \\
\graymidrule
$\mathbf{R_3}$ & ``\mytt{The following scenes appear in the video, not necessarily in this order:} [comma-separated caption list]\mytt{. Of those scenes, which occurs} [either first or last] \mytt{?}'', followed by a number of different scene captions from the pseudovideo. \\
\graymidrule
$\mathbf{R_4}$ & ``\mytt{One of the scenes in the video can be described as "}[caption]\mytt{". Describe the scene immediately} [either ``\mytt{before}'' or ``\mytt{after}''] \mytt{it.}'', followed by a number of different scene captions from the pseudo video, or an option stating that the given scene is the first/last one and hence has no scene before/after it respectively.
This option may either be correct or wrong. \\
\midrule
$\mathbf{A_1}$ & ``{\mytt{How many different scenes appear in the video?}}'', followed by a number of numerical answers. \\
\graymidrule
$\mathbf{A_2}$ & ``\mytt{There are }[number of scenes]\mytt{ in the video. What does scene }[number between 1 and the number of scenes]\mytt{ depict?}'', followed by a number of different scene captions from the pseudo video. \\
\bottomrule
\end{tabularx}
\caption{Descriptions and abbreviations for all questions we experiment with.
The $\mathbf{R}_i$ questions focus on asking about relative scene orderings, whereas the $\mathbf{A}_i$ variants focus on localizing scenes in an absolute fashion within a pseudo video.}
\label{tab:question-table}
\end{table*}

We can group the examined questions in two broad categories depending on the type of temporal reasoning required to answer them:
\begin{itemize}
    \item Questions for which an understanding of \emph{relative} temporal concepts is needed.
    Examples include asking the model to choose the option that describes the order of scenes in the video, or to determine whether a scene with a certain description appears before or after another.
    \item Questions for which the model must be able to reason about \emph{absolute} quantities.
    This may be in the form of absolute temporal localization, \eg providing a caption for the $i$-th scene, or in the form of counting, \eg counting the number of different scenes that appear in the video.
\end{itemize}

In principle, both of these question types target capabilities that would be desirable in a video-language model. 
Table \ref{tab:question-table} lists six different questions with which we experiment, along with their corresponding abbreviations.
Of these questions, $\mathbf{R_1}, \mathbf{R_2}, \mathbf{R_3}, \mathbf{R_4}$ can be characterized as requiring relative temporal reasoning, and $\mathbf{A_1}, \mathbf{A_2}$ as requiring absolute temporal reasoning.

\begin{table}[t]
    \centering
    \begin{tabularx}{0.735\linewidth}{lY}
        \toprule
        Task name & Abbreviation \\
        \midrule
        Action Count & AC \\
        Object Count & OC \\
        Action Sequence & AS \\
        Object Shuffle & OS \\
        Scene Transition & ST \\
        Action Localization & AL \\
        Action Antonym & AA \\
        Unexpected Action & UA \\
        Egocentric Sequence & ES \\
        Moving Direction & MD \\
        \bottomrule
    \end{tabularx}
    \caption{TVBench \cite{cores2025losttimenewtemporal} task name abbreviations.}
    \label{tab:task_abbrevs}
\end{table}
\section{Experiments}

\label{sec:experiments}

The Video LLMs we use are based on the recent LongVU \cite{shen2024longvu} architecture.
LongVU utilizes DINOv2 \cite{oquab2023dinov2} feature-space similarity to select a subset of the input frame sequence, and then applies additional heuristics to represent each remaining frame in high or low resolution.
Our choice is based on the strong performance of these models on various benchmarks, the availability of training code, and checkpoints for both image- and video-trained models with Llama3.2 \cite{grattafiori2024llama3} and Qwen2 \cite{bai2025qwen2} backbones.

Unless mentioned otherwise, we use a batch size of 64, as in the original paper, and a slightly modified learning rate schedule, which led to somewhat faster convergence in preliminary experiments.
The learning rate increases linearly  to $5 \cdot 10^{-6}$ for the first 3\% of the total training steps and then follows a cosine decay schedule for the remainder of training.
All hyperparameters specific to the LongVU algorithm are kept the same as in the original work \cite{shen2024longvu}.

As discussed, our benchmark of choice is TVBench, due to its focus on temporal aspects of video understanding.
In the following, we will refer to TVBench tasks using abbreviations listed in Table \ref{tab:task_abbrevs}.
For more details on the nature of the tasks, we refer the reader to the original publication \cite{cores2025losttimenewtemporal}.

\begin{table*}[ht]
\centering{}
\setlength{\tabcolsep}{2.0pt}

\begin{tabularx}{\textwidth}{lcccYYYYYYYYYYcY}
\toprule 
Configuration &~~& LLM &~~& AC & OC & AS & OS & ST & AL & AA & UA & ES & MD &~~& Avg. \\
\midrule
Chance level && && 25.0 & 25.0 & 50.0 & 33.3 & 50.0 & 25.0 & 50.0 & 25.0 & 25.0 & 25.0 && 33.3 \\
\midrule
Video-trained && Llama3.2-3B && 31.0 & 59.4 & 70.2 & 33.3 & 76.2 & 44.4 & 59.1 & 32.9 & 27.0 & 75.9 && \textbf{51.2} \\
\downrightarrow shuffled frame evaluation && Llama3.2-3B && 30.4 & 32.4 &  54.5 & 33.8 & 54.6 & 28.7 & 51.9 & 31.7 & 25.0 & 28.4 && 38.8 \\
Image-trained (1 pseudo video step) && Llama3.2-3B && 27.1 & 45.3 & 60.2 & 32.4 & 54.6 & 36.2 & 55.0 & 35.4 & 30.0 & 77.6 && 45.6 \\
\downrightarrow shuffled frame evaluation && Llama3.2-3B && 27.6 & 33.1 & 56.3 & 32.9 & 51.3 & 29.4 & 52.5 & 36.6 & 31.5 & 26.3 && 38.8 \\

\midrule
Video-trained && Qwen2-7B && 33.9 & 61.5 & 75.3 & 36.9 & 78.4 & 60.0 & 65.0 & 34.1 & 31.5 & 78.9 && \textbf{55.8} \\
\downrightarrow shuffled frame evaluation && Qwen2-7B && 31.7 &  37.2 &  54.2 & 38.2 & 58.4 & 30.0 & 50.0 & 28.0 & 30.0 & 24.1 && 39.7 \\
Image-trained  (1 pseudo video step) && Qwen2-7B && 31.0 &  53.4 &  66.8 & 37.8 & 75.1 & 49.4 & 58.1 & 36.6 & 32.5 & 66.8 && 50.5 \\
\downrightarrow shuffled frame evaluation && Qwen2-7B && 31.1 &  33.8 &  55.6 & 37.3 & 51.9 & 32.5 & 51.2 & 36.6 & 30.5 & 27.6 && 40.0 \\

\bottomrule 
\end{tabularx}

\caption{Comparison of image- and video-trained LLM performance on TVBench \cite{cores2025losttimenewtemporal}. Image-trained LLMs perform significantly above chance level for both Llama3.2 and Qwen2. The drop in performance when video frames are shuffled indicates that all models do learn some form of temporal reasoning, even those trained only on images.
}
\label{tab:image_trained_llms}
\end{table*}

\subsection{Temporal reasoning in image-trained LLMs}

We first conduct experiments to determine how well the two LongVU models that have only been trained on images perform on TVBench.
Ideally, this would be done by simply evaluating the provided checkpoints on the benchmark.
However, when changing from image input to video input, the manner in which images are tokenized changes slightly.
We thus find that
directly evaluating the image-pretrained checkpoint on TVBench results in the model outputting text that does not follow the multiple-choice format required.
This issue can be resolved by training the models for as little as one step on samples consisting of multiple frames.We therefore present results when training for one step on pseudo videos  with question $\mathbf{R_1}$.
Given the short duration of training and the fact that these models never see real videos, we deem these configurations to be sufficiently close to LLMs that have only been trained on images and thus use them as their proxy for our evaluation.

Table~\ref{tab:image_trained_llms} shows the achieved performance of image-trained models and compares these to the full video-trained LongVU models.
When trained only on text and images, both Llama and Qwen score significantly above chance level, despite the models never having been trained on videos.
The gap between image-trained models and video-trained models is surprisingly small ($\sim 5\%$) when considering the additional data and compute required for training.
Note, also, that providing the frames in a shuffled order results in lower performance for both models.
Although the resulting accuracy is not on chance level, it is consistent with what is reported in the same experimental setting for several models in the original work \cite{cores2025losttimenewtemporal}.
This indicates that the models are indeed performing temporal reasoning to some extent and not just exploiting potential weaknesses or ``shortcuts'' of the dataset.

Further interesting observations can be made when focusing on some of the subtasks where the models perform the best.
Two of these, Object Count (OC) and Moving Direction (MD), are based on videos sourced from the CLEVRER \cite{yi2019clevrer} dataset.
Although the models have not seen videos from this dataset, the training recipe does include images from the CLEVR dataset \cite{johnson2017clevr} which are visually similar to CLEVRER.
These results suggest that the LLM can, to some extent, solve temporal tasks simply by training on visually similar image-text data.
To gain further insights into this matter, one could retrain image-text models while excluding CLEVR-source images and questions, and assess the impact on the OC and MD subsets.
The scale of this experiment, however, exceeds our computational resources.

Overall, the results suggest that Video LLMs are capable of learning simple temporal reasoning purely from text and images, without ever being trained on videos.
Understanding the mechanisms involved would provide valuable insights regarding the way these models process video, and we encourage future research in this direction. 
Furthermore, these results raise the question whether text-image data could be used to further improve the temporal reasoning capabilities of models.
The pseudo-video setup with which we experiment in the next section is an example how this could be achieved, but we anticipate that there are various other effective experimental setups.

\subsection{Pseudo video-based training}

\begin{table*}[ht]
\centering{}
\setlength{\tabcolsep}{5.0pt}

\begin{tabularx}{\textwidth}{lcccYYYYYYYYYYcY}
\toprule 
Configuration && LLM && AC & OC & AS & OS & ST & AL & AA & UA & ES & MD && Avg. \\
\midrule
Chance level &&&& 25.0 & 25.0 & 50.0 & 33.3 & 50.0 & 25.0 & 50.0 & 25.0 & 25.0 & 25.0 && 33.3 \\
\midrule
Videos && Llama3.2-3B&& 31.0 & 59.4 & 70.2 & 33.3 & 76.2 & 44.4 & 59.1 & 32.9 & 27.0 & 75.9 && 51.2 \\
Pseudo videos && Llama3.2-3B&& 31.7 & 54.1 & 75.1 & 38.7 & 84.9 & 41.9 & 57.2 & 34.1 & 28.5 & 83.6 && \textbf{53.5} \\
Videos + pseudo videos && Llama3.2-3B && 33.0 & 53.4 & 75.3 & 36.4 & 83.8 & 49.4 & 55.0 & 30.5 & 28.0 & 82.3 && 53.4 \\

\midrule
Videos && Qwen2-7B && 33.9 & 61.5 & 75.3 & 36.9 & 78.4 & 60.0 & 65.0 & 34.1 & 31.5 & 78.9 && \textbf{55.8} \\
Pseudo videos && Qwen2-7B && 32.6 &  53.4 & 74.8 & 39.1 & 80.0 & 50.0 & 60.0 & 35.4 & 28.5 & 58.2 && 51.9 \\
Videos + pseudo videos && Qwen2-7B && 32.6 &  57.4 &  75.3 & 38.7 & 82.7 & 52.5 & 63.8 & 35.4 & 27.0 & 74.6 && 54.4 \\

\bottomrule 
\end{tabularx}

\caption{Results achieved by the best configurations for our pseudo-video-trained LLMs. Surprisingly, the results are close to, or even higher than those of video-trained LLMs, raising the question how well real videos are utilized during the training process. Finetuning the video-trained LLM with pseudo videos yields mixed results: performance on some tasks improves, whereas on others it slightly degrades.}
\label{tab:pseudovideo_best}
\end{table*}

\begin{table*}[ht]
\centering
\setlength{\tabcolsep}{5.0pt}

\begin{tabularx}{0.75\textwidth}{lcYYYYYYYYYYcY}
\toprule 
Configuration && AC & OC & AS & OS & ST & AL & AA & UA & ES & MD && Avg. \\
\midrule
Chance level && 25.0 & 25.0 & 50.0 & 33.3 & 50.0 & 25.0 & 50.0 & 25.0 & 25.0 & 25.0 && 33.3 \\
\midrule
$\mathbf{R_1}, S = 4$ && 26.9 & 52.7 & 72.1 & 36.9 & 78.9 & 34.4 & 54.7 & 30.5 & 28.0 & 82.8 && 50.3 \\
$\mathbf{R_2}, S = 4$ && 28.0 & 53.4 & 72.5 & 38.7 & 62.2 & 35.6 & 56.2 & 30.5 & 30.0 & 84.1 && 50.1 \\
$\mathbf{R_3}, S = 6$ && 31.5 & 50.0 & 75.7 & 28.9 & 76.8 & 41.2 & 55.6 & 24.4 & 28.5 & 84.5 && 51.4 \\
$\mathbf{R_4}, S = 6$ && 29.7 & 52.7 & 74.1 & 39.1 & 64.3 & 39.4 & 55.9 & 42.7 & 30.0 & 85.8 && 51.6 \\
$\mathbf{A_1}, S = 6$ && 20.3 & 45.3 & 68.6 & 33.3 & 61.1 & 33.8 & 59.4 & 41.5 & 28.5 & 79.7 && 46.9 \\
$\mathbf{A_2}, S = 6$ && 28.2 & 44.6 & 73.7 & 32.9 & 64.9 & 39.4 & 56.2 & 42.7 & 28.0 & 78.9 && 49.5 \\

\bottomrule 
\end{tabularx}

\caption{Effect of different questions used during training. Overall, relative questions perform better than absolute ones.}
\label{tab:various_questions}
\end{table*}

\begin{table*}[ht]
\centering

\begin{tabularx}{0.9\textwidth}{lcYYYYYYYYYYcY}
\toprule 
Configuration && AC & OC & AS & OS & ST & AL & AA & UA & ES & MD && Avg. \\
\midrule
Chance level && 25.0 & 25.0 & 50.0 & 33.3 & 50.0 & 25.0 & 50.0 & 25.0 & 25.0 & 25.0 && 33.3 \\
\midrule
$\mathbf{R_1}, F = 1$ && 25.9 & 50.0 & 68.6 & 35.1 & 63.2 & 32.5 & 59.1 & 24.4 & 29.0 & 73.3 && 47.4 \\
$\mathbf{R_1}, F = 5$ && 26.9 & 52.7 & 72.1 & 36.9 & 78.9 & 34.4 & 54.7 & 30.5 & 28.0 & 82.3 && 50.3 \\
$\mathbf{R_1}, F = 10$ && 27.8 & 50.7 & 73.2 & 36.4 & 78.4 & 40.0 & 55.9 & 35.4 & 27.5 & 84.1 && 51.2 \\
$\mathbf{R_1}, F = 20$ && 27.1 & 48.6 & 73.5 & 40.9 & 83.8 & 48.8 & 58.4 & 34.1 & 24.0 & 82.3 && \textbf{52.2} \\
$\mathbf{R_1}, F = 40$ && 30.0 & 54.1 & 73.9 & 40.4 & 82.2 & 41.9 & 56.2 & 29.3 & 24.0 & 81.9 && 52.1 \\
\graymidrule
$\mathbf{R_3}, F = 5$ && 31.5 & 50.0 & 75.7 & 28.9 & 76.8 & 41.2 & 55.6 & 24.4 & 28.5 & 84.5 && 51.4 \\
$\mathbf{R_3}, F = 20$ && 31.9 & 57.4 & 77.3 & 36.0 & 71.9 & 47.5 & 54.4 & 28.0 & 28.5 & 75.0 && 52.0 \\
\graymidrule
$\mathbf{R_4}, F = 5$ && 29.7 & 52.7 & 74.1 & 39.1 & 64.3 & 39.4 & 55.9 & 42.7 & 30.0 & 85.8 && 51.6 \\
$\mathbf{R_4}, F = 20$ && 27.6 & 50.0 & 71.8 & 31.5 & 64.3 & 43.7 & 57.5 & 35.4 & 28.0 & 71.1 && 48.7 \\

\bottomrule 
\end{tabularx}

\caption{Effects of varying the maximum number of frames per scene.
The optimal number of frames and robustness to this hyperparameter is somewhat question-dependent. However, surprisingly high scores can be reached with all of the relative temporal reasoning questions.}
\label{tab:num_frames}
\end{table*}

\begin{table*}[ht]
\centering{}

\begin{tabularx}{0.9\textwidth}{lcYYYYYYYYYYcY}
\toprule 
Configuration && AC & OC & AS & OS & ST & AL & AA & UA & ES & MD && Avg. \\
\midrule
Chance level && 25.0 & 25.0 & 50.0 & 33.3 & 50.0 & 25.0 & 50.0 & 25.0 & 25.0 & 25.0 && 33.3 \\
\midrule
$S = 2, F = 5$ && 28.5 & 49.3 & 72.3 & 34.2 & 60.0 & 35.0 & 57.5 & 28.0 & 33.5 & 80.2 && 49.3 \\
$S = 2, F = 10$ && 26.9 & 49.3 & 72.3 & 31.5 & 60.0 & 35.6 & 57.5 & 25.6 & 31.0 & 81.5 && 48.6 \\
$S = 4, F = 5$ && 26.9 & 52.7 & 72.1 & 36.9 & 78.9 & 34.4 & 54.7 & 30.5 & 28.0 & 82.3 && \textbf{50.3} \\
$S = 8, F = 5$ && 27.1 & 49.3 & 73.2 & 35.5 & 77.3 & 36.2 & 57.8 & 25.6 & 23.5 & 82.7 && 50.0 \\

\bottomrule 
\end{tabularx}

\caption{Effects of varying the maximum number of scenes for question $\mathbf{R_1}$.
Providing more than two scenes is particularly important for the ``Scene Transition'' task, but there is no one specific setting that performs equally well for all tasks.}
\label{tab:num_scenes}
\end{table*}

We use COCO \cite{lin2014mscoco} as the source of annotated images for our pseudo-video setup.
COCO captions are relatively short, which may be beneficial when creating questions focusing on temporal reasoning rather than detailed scene descriptions.
Furthermore, the LLaVA-OneVision data~\cite{li2024llavaonevison}, on which the LongVU models have been trained, includes COCO images, and this may make it easier for the models to utilize already acquired image knowledge.

When creating pseudo videos as described in Section \ref{sec:method}, we will refer to the maximum number of scenes (\ie, number of distinct images sampled) as $S$ and to the maximum number of frames per scene (\ie, repetitions of an image to which we apply an affine transformation) as $F$.
The number of scenes is sampled uniformly at random from $\{1, 2, \ldots, S\}$, except in cases where a single scene would render the question trivial, \eg $\mathbf{R_1}$, in which case we sample from $\{2, 3, \ldots, S\}$ instead.
The number of frames per scene is sampled uniformly at random from $\{1, 2, \ldots, F\}$.

For question creation, we typically use 3 wrong options, except for $\mathbf{R_2}$ where the number of wrong options is always 1, and for $\mathbf{R_3}$, where we make this equal to the length of the list of captions, which is sampled from $\{1, 2, 3\}$.
Unless stated otherwise, all of the models presented were trained for 2 epochs on 100,000 pseudo videos, corresponding to  3,125 training steps.

\PAR{Pseudo-video-trained TVBench results.}
In \Cref{tab:pseudovideo_best} we present the results we achieved when finetuning image- and video-trained LongVU checkpoints with our pseudo video scheme.
Here, we report the results of the best pseudo-video configuration, as found empirically through experiments.
For both Llama3.2 checkpoints, these were achieved with $(\mathbf{R_1}: S=4, F=20)$ and $(\mathbf{R_3}: S=6, F=5)$ as the question generation setup.
For the two Qwen2 models, the best results were achieved with $(\mathbf{R_1}: S = 4, F = 5)$, and we observed marginally worse (-0.4) results for $F = 10$.
Because we use A40s compared to H100s used in the original LongVU work \cite{shen2024longvu}, our GPU memory is insufficient for training Qwen2 with $F > 10$, and thus we could not test setups with 20 frames like the one used for LLama3.2.
The experiments finetuning video-trained models on pseudo videos aim to determine whether this type of training is beneficial for LLMs that are also trained on real videos.
Ideally, this would be done by including pseudo-video training as an intermediate phase between image and real video training.
However, this would mean training the LongVU checkpoints on the video data originally used. This was challenging in terms of gathering the datasets used, as well as computationally prohibitive given our resources.
We are thus only able to test the influence of pseudo videos when these are used in a training phase \emph{after} training on real videos.

We observe that the pseudo-video-trained Llama model (53.5\%) outperforms its video-trained counterpart (51.2\%).
The tasks that appear to particularly benefit from this type of training are ``Action Sequence'' (AS), ``Scene Transition'' (ST) and ``Moving Direction'' (MD).
This is perhaps intuitive, as these tasks heavily rely on the model's ability to relate frames over time.
Finetuning the video-trained version yields a slight increase on average, with the result (53.4\%) approximately matching the performance achieved by finetuning the image-trained model.
We also note, however, that the tasks ``Object Count'' (OC) and ``Action Antonym'' (AA) are negatively impacted by this additional finetuning, perhaps indicating some amount of task-specific forgetting.

For the Qwen2 model, the results indicate that pseudo-video training is less effective.
This can partially be attributed to the fact that the image-trained baseline for Qwen2 is already quite capable in the ST and AS tasks, and thus does not benefit as much.
Furthermore, it is noteworthy that the MD performance gets \textit{worse} compared to the image baseline, whereas it got better for the Llama-based model, while these models are trained on the same image and pseudo-video data. 
We encourage future work to further explore such differences between LLMs.
For the video-trained Qwen2 model, despite the improved performance in ``Scene Transition'', we were unable to find pseudo-video setups that maintain or increase average accuracy.
Although the results are in general not as surprising, the best pseudo-video-trained Qwen2 model scores 51.9\% average accuracy, which is only 4\% lower than the video-trained counterpart.
Given that the video model was trained on approximately 550,000 real videos \cite{shen2024longvu}, this performance gap is still not as large as one would expect from such a simple baseline.
In terms of further exploring pseudo-video training for this model, we note that the data generation process is flexible: if, for example, we wanted the model to learn about \emph{reappearances} of scenes, this constraint could easily be incorporated in the sampling process described in Section \ref{sec:method}.
Another approach could be utilizing existing \emph{mask annotations}: pseudo videos that feature objects moving in various patterns open up an entire new domain of questions, and it could be that this particular model has more to gain from this type of spatiotemporal reasoning.

While the benchmark is far from solved, these results suggest that the current training procedures for Video LLMs may not be as effective in learning useful temporal features from real videos as we would expect.
Specifically, we are able to get results close to---or in the case of Llama3.2 even slightly better than---the ones achieved by video-trained LLMs using pseudo videos that definitely do not capture the complexity of real videos.
We would therefore expect video-trained LLMs that had access to realistic video dynamics to perform significantly better, yet this does not appear to be the case.
Considering the amount of annotation effort, storage requirements and computing resources spent on training LLMs on large video datasets, we believe it is important to identify and address the bottlenecks that result in this simple baseline currently being so competitive.

As for using pseudo videos as an additional training phase, the results suggest that video-trained models may not necessarily benefit, and that even when they benefit on average, task-specific performance may still decrease.
The impact of this type of training as an intermediate instead of as the last training phase remains an open question.

Next, we investigate the hyperparameters for pseudo video generation and the type of questions used in training.
Unless stated otherwise, \emph{these experiments are conducted on the Llama3.2 model}, since training is significantly faster.

\begin{table*}[ht]
\centering{}

\begin{tabularx}{\textwidth}{lcYYYYYYYYYYcY}
\toprule 
Configuration && AC & OC & AS & OS & ST & AL & AA & UA & ES & MD && Avg. \\
\midrule
Chance level && 25.0 & 25.0 & 50.0 & 33.3 & 50.0 & 25.0 & 50.0 & 25.0 & 25.0 & 25.0 && 33.3 \\
\midrule
$N = \textrm{10,000}$; 312 steps && 26.5 & 48.6 & 72.1 & 32.9 & 68.1 & 32.5 & 57.8 & 30.5 & 28.5 & 84.1 && 49.2 \\
$N = \textrm{100,000}$; 3,125 steps && 26.9 & 52.7 & 72.1 & 36.9 & 78.9 & 34.4 & 54.7 & 30.5 & 28.0 & 82.3 && \textbf{50.3} \\
$N = \textrm{500,000}$; 15,625 steps && 28.4 & 50.7 & 72.3 & 36.9 & 82.2 & 34.4 & 55.6 & 31.7 & 21.5 & 64.7 && 48.7 \\

\bottomrule 
\end{tabularx}

\caption{Effects of varying the maximum number of pseudo videos $N$ when training with question $\mathbf{R_1}$. Again there seem to be task-specific optima, but especially for the ``Egocentric Sequence'' and ``Moving Direction'' tasks training with more data is significantly worse.}
\label{tab:num_samples}
\end{table*}

\PAR{Questions posed to the model.}
To ablate the effect of each question in Table~\ref{tab:question-table}, we train separate models on each of them in isolation.
We set $F = 5$ and vary the number of scenes $S$ depending on the question, since in some cases it impacts the diversity of possible answers more significantly.
For example, question $\mathbf{A_1}$ with $S = 4$ and four candidate answers would always present the model with a permutation of 1, 2, 3, 4, thus limiting training sample variety.
Therefore, for some questions we use a slightly higher $S = 6$.

From the corresponding results in Table \ref{tab:various_questions}, we tentatively conclude that querying the Video LLM with questions that require some kind of \textit{absolute} temporal reasoning is less effective.
This could either be a phenomenon specific to pseudo-video training, or a more general characteristic of the models being used.
Determining which is the case would require a study of similar questions paired with real videos.
Another way of investigating this further would be the construction of a video-based benchmark that features tasks targeting each of these reasoning types separately.
This would aid in determining whether Video LLMs find absolute temporal reasoning more challenging during inference as well.

\PAR{Number of frames per scene.}
The number of frames per scene partially determines the length of the created pseudo videos.
For this experiment, we focus on $\mathbf{R_1}$, $\mathbf{R_3}$, and $\mathbf{R_4}$ due to encouraging results in \Cref{tab:various_questions}.
In Table \ref{tab:num_frames}, we observe that, with $\mathbf{R_1}$, the performance improves until $F = 20$, at which point it decreases slightly.
The largest improvement is for $F = 5$: compared to $F = 1$, this allows a scene to last for more than one frame (on average, 3), perhaps bringing our pseudo videos closer to real ones compared to a literal sequence of a number of completely different images.
Considering the other question types, $\mathbf{R_3}$ seems to benefit as well, although less so in ``Scene Transition'' and ``Moving Direction''.
$\mathbf{R_4}$, however, exhibits a significant performance drop with more frames.
We also report that the mix of $(\mathbf{R_1}: F = 20, S = 4)$ and $(\mathbf{R_3}: F = 20, S = 6)$ achieves 51.5\% as opposed to the best result in Table \ref{tab:pseudovideo_best} (53.5\%), which used $F = 5$ for $\mathbf{R_3}$.
From this we conclude that the optimal $F$ value varies between questions and combinations of questions.

\PAR{Number of scenes per pseudo video.}
Besides influencing pseudo-video length, the number of scenes impacts the complexity of the videos and the questions more than $F$ does, \eg, when setting $S = 8$ in question $\mathbf{R_1}$, the model is asked to place up to 8 scenes in the correct order of appearance.
We study the effect of changing $S$ for $\mathbf{R_1}$ by fixing $F = 5$, except for $S = 2$ where we also try $F = 10$.
We do this to rule out performance differences caused by $S = 2, F = 5$ yielding a smaller average and maximum number of frames (max. 10) compared to $S = 4, F = 5$ (max. 20).
Table \ref{tab:num_scenes} shows a meaningful performance improvement from $S = 2$ to $S = 4$, particularly for ``Scene Transition'', while there is a marginal decrease from $S = 4$ to $S = 8$.
We conclude that the model learns more than just a simple ordering of two different scenes, and that this knowledge is particularly useful for the ``Scene Transition'' task.

\PAR{Training dataset size.}
Here, we use $\mathbf{R_1}$ with $S = 4, F = 5$ and train for the equivalent of two epochs while varying the number of generated pseudo videos.
From the results in Table \ref{tab:num_samples}, we conclude that training on more than 100,000 pseudo videos improves performance on the ``Scene Transition'' task, which is similar to the question posed during training.
However, the overall performance decreases, suggesting that the model begins to overfit on a specific task.
In addition to these results, we note that in earlier experimentation rounds---albeit under less principled setups---we never observed stable performance improvements beyond approximately 3,000 steps/100,000 pseudo videos.

\vspace{-5pt}
\section{Conclusion}

\vspace{-4pt}
We presented experimental results that cast doubt on the effectiveness of current training schemes for Video LLMs.
First, we showed that two models based on the recent LongVU approach trained on just images and text are surprisingly effective at temporal reasoning.
Without ever training on real videos, these models achieve accuracies significantly above chance level on the recent TVBench benchmark, which evaluates various aspects of temporal reasoning in videos.
Second, we investigated a simple setup where these models are finetuned on pseudo videos constructed by concatenating and perturbing COCO images.
When trained with certain questions that require reasoning about the relative order of scenes in pseudo videos, we observe performance improvements comparable to or even higher than those achieved by finetuning on real videos.

The small gap between video-trained Video LLMs and our baselines indicates that better utilization of image datasets during training may reduce the need for videos.
It also suggests that real videos, though more information-rich, may be contributing less than expected to temporal understanding in Video LLMs.
This points to a bottleneck in the current paradigm, either due to the quality of video and captioning data, the training setup, or architectural weaknesses.
Since training Video LLMs becomes increasingly expensive with larger models and datasets, investigating this bottleneck is crucial for efficient utilization of computational resources.

\PAR{Future Directions.}  Experiments involving more video benchmarks and other Video LLM architectures may further cement our findings. 
Comparing pseudo-video to video training would be particularly interesting on tasks that specifically require understanding of complex video dynamics, as well as in long-range settings involving potentially hours-long videos.

\PAR{Acknowledgments.} This project is partially funded by BMBF projects NeuroSys-D (\verb|03ZU1106DA|) and 6GEM (\verb|16KISK036K|).
Compute resources were granted by the Gauss Centre for Supercomputing e.V. through the John von Neumann Institute for Computing on the GCS Supercomputer JUWELS \cite{JUWELS} at Julich Supercomputing Centre. 

{
    \small
    \bibliographystyle{ieeenat_fullname}
    \bibliography{main}
}

\end{document}